\documentclass{bmvc2k}
\usepackage[ruled,vlined]{algorithm2e}
\usepackage{multirow}
\usepackage{amsmath}
\usepackage{url}
\title{Paying more Attention to Snapshots of Iterative Pruning: Improving Model Compression via Ensemble Distillation}

\addauthor{Duong H. Le}{1610580@hcmut.edu.vn}{1}
\addauthor{Trung-Nhan Vo}{1612372@hcmut.edu.vn}{1}
\addauthor{Nam Thoai}{namthoai@hcmut.edu.vn}{1}
\addinstitution{
  Ho Chi Minh City University of Technology \\
  Ho Chi Minh, Vietnam
}

\runninghead{Le et al}{Pay Attention to Snapshots of Pruning}

\def\eg{\emph{e.g}\bmvaOneDot}

\def\etal{\emph{et al}\bmvaOneDot}

\begin{document}

\maketitle

\begin{abstract}
  Network pruning is one of the most dominant methods for reducing the heavy inference cost of deep neural networks. Existing methods often iteratively prune networks to attain high compression ratio without incurring significant loss in performance. However, we argue that conventional methods for retraining pruned networks (i.e., using small, fixed learning rate) are inadequate as they completely ignore the benefits from snapshots of iterative pruning. In this work, we show that strong ensembles can be constructed from snapshots of iterative pruning, which achieve competitive performance and vary in network structure.  Furthermore, we present a simple, general and effective pipeline that generates strong ensembles of networks during pruning with \textit{large learning rate restarting}, and utilizes knowledge distillation with those ensembles to improve the predictive power of compact models. In standard image classification benchmarks such as CIFAR and Tiny-Imagenet, we advance state-of-the-art pruning ratio of structured pruning by integrating simple $\ell_1$-norm filters pruning into our pipeline. Specifically, we reduce 75-80\% of total parameters and 65-70\% MACs of numerous variants of ResNet architectures while having comparable or better performance than that of original networks. Code is available at \url{https://github.com/lehduong/kesi}.
\end{abstract}

%------------------------------------------------------------------------- 
\section{Introduction}
\label{sec:intro}

\textbf{Motivation} Researchers have extensively exploited deep and wide networks for the sake of achieving superior performance on various tasks. Most of state-of-the-art networks are extremely computationally expensive and require excessive memory. However, real-world applications usually require running deep neural networks on edge devices for various reasons: user privacy, security, real-time analysis, offline capability, reducing cost for server deployment, and so on. Adopting large and cumbersome networks to such resource-constrained environments is challenging due to restrictions of memory, computational power, energy consumption, and so on.
\paragraph{Background}
Network pruning \cite{lecun1990optimal, reed1993pruning, han2015learning, li2016pruning} reduces a cumbersome and over-parameterized network to compact one by removing unnecessary weights and connections of networks. It is widely believed that small networks pruned from large, over-parameterized networks achieve superior performance than those trained from scratch \cite{frankle2018lottery, renda2020comparing, li2016pruning, luo2017thinet}. A plausible explanation to this phenomenon is the lottery ticket hypothesis \cite{frankle2018lottery} i.e. large, over-parameterized networks contain many optimal sub-networks i.e. winning tickets. In particular, network pruning could be done in two manners:  \textit{one-shot pruning} - prune a network with the desired compression ratio and retrain it only \textbf{one} time, or \textit{iterative pruning} - only prune small ratio of the original network, retrain and repeat that process until the target size is reached. It has been shown that iterative pruning could lead to a greater compression ratio compare to one-shot pruning approaches \cite{han2015learning, luo2017thinet, li2016pruning, renda2020comparing}. Furthermore, Frankle \etal \cite{frankle2018lottery} point out that iteratively-pruned-winning-tickets learn faster and reach higher test accuracy at smaller network size.

On the other hand, ensembles of neural networks are known to be much more robust and accurate than individual networks \cite{huang2017snapshot, ashukha2020pitfalls, snoek2019can}. In spite of their superior performance, the tremendous cost of training and inference of ensembles makes them less attractive in practice. For the purpose of accelerating training time of ensembles, prior works proposed methods encouraging models to converge to different local minimums during training  \cite{huang2017snapshot, garipov2018loss, yang2019snapshot}. To reduce inference time of ensembles, one could use a single network to mimic behavior of ensembles as pioneered by born-again tree \cite{breiman1996born} and knowledge distillation \cite{hinton2015distilling, balan2015bayesian,bucilua2006model, malinin2019ensemble}. In above approaches, although small networks can not achieve comparable performance with ensembles of networks, \textit{dark knowledge} transferred from teachers to student network could bridge the gap between their predictive powers.

\paragraph{Our proposal} While existing methods of iterative pruning are more effective than one-shot pruning, the snapshots at each pruning iteration are mostly overlooked. We consider leveraging the snapshots of iterative pruning to take the performance of compact models to the next level.

In this work, we propose a simple pipeline for model compression by slightly modifying the standard approach. Specifically, we make use of \textit{large learning rate restarting} at each pruning iteration to retrain pruned networks. Hence, each retraining step could be considered as a \textit{cycle} of \textit{Snapshot ensemble} \cite{huang2017snapshot}. Utilizing both large learning rate restarting and pruning foster the diversity between snapshots, thus, constructing strong ensembles. Once achieved the desired compression ratio, we then distill the knowledge from the ensembles of snapshots of iterative pruning to the final model. Our method acquires the advantages of network pruning, ensembles learning, and knowledge distillation. To the best of our knowledge, this is the first work attempting to exploit snapshots of iterative pruning to further improve the performance of pruned networks. 

\paragraph{Our main contributions} The contributions of our work are summarized as below:
\begin{enumerate}
  \item We empirically show that fine-tuning with large learning rate restarting can achieve competitive or better results than the common strategy (i.e. small, fixed learning rate) on a range of standard datasets and architectures. Surprisingly, such simple modification can create very strong baselines for both structured and unstructured pruning.
  \item We demonstrate that snapshots of iterative pruning could construct strong ensembles.
  \item We propose a simple pipeline to combine knowledge distillation from ensembles and iterative pruning. We empirically show that our approach can achieve state-of-the-art pruning ratio by reducing $75-80\%$ of parameters and $65-70\%$ MACs on numerous variants of ResNet while having comparable or better results than original networks.
\end{enumerate}
%The paper is organized as follow: Section 2 outlines some effort has been done in the field of model compression. In section 3, we provide background on knowledge distillation. Section 4, we propose our finding of constructing ensembles from snapshots of iterative pruning. Section 5 presents our pipeline for effective compress neural networks. In section 6, we discuss the emperical results of our proposed methods and ablation study. Finally, section 7 draws conclusion and talks about promising directions for future works.

%------------------------------------------------------------------------- 
% RELATED WORK
\section{Related Work}
\paragraph{Knowledge Distillation}  The approach of training small, efficient student network to mimic behavior of large, over-parameterized network has been proposed for a long time \cite{bucilua2006model} and was recently repopularized in \cite{hinton2015distilling, ba2014deep}. Later, knowledge distillation was extended to various aspects, transferring knowledge from intermediate layers \cite{romero2014fitnets, zagoruyko2016paying}, allowing teachers and students to guide each others \cite{zhang2018deep}, using teacher and student with the same architecture \cite{furlanello2018born, yang2019snapshot, yang2019training, bagherinezhad2018label}, distilling knowledge in multiple steps \cite{mirzadeh2019improved}. To address the cost of training two networks in knowledge distillation, \cite{zhu2018knowledge, zhang2018deep, yang2019snapshot} propose online approaches to train the student and teacher networks in one generation. Furthermore, Anil \etal \cite{anil2018large} adopt knowledge distillation to accelerate the training of large scale neural networks. Universally Slimmable networks \cite{yu2019universally} provide an ensemble of sub-networks that has implicit knowledge distillation through shared weights.
\paragraph{Network Pruning} The idea behind network pruning is to reduce the redundant weights and connections of original network to achieve compact networks without losing much performance \cite{han2015learning, li2016pruning}. In general, pruning can be divided into two categories: structured pruning and unstructured pruning. Unstructured pruning \cite{hanson1989comparing, lecun1990optimal, han2015learning, srinivas2015data, guo2016dynamic} always results in sparse weight matrices, which can not directly accelerate the inference efficiency without specialized hardware/libraries. In contrast, structured pruning approaches \cite{li2016pruning, he2017channel, yu2018nisp, lin2019towards, molchanov2016pruning} remove the redundant weights at the level of filters/channels/layers, thus, speeding up the inference of networks directly. There are numerous approaches to determine redundant filters/weights: \cite{luo2017thinet} use statistic information of the next filters to select unimportant filters, \cite{li2016pruning} prune the filters that have smallest norms in each layer, \cite{molchanov2016pruning} select the filters to minimize the construction loss estimated with Taylor expansion. As these criteria are rough estimations of weight's importance, pruning a large number of filters/weights at once might break down and lead to inferior performance compare to iterative pruning \cite{han2015learning, li2016pruning}. Recently, Liu \etal \cite{liu2018rethinking} empirically show that training the pruned model from scratch can also achieve comparable or even better performance than fine-tuning. While the efficacy of network pruning remains an open question, in this work, we propose exploiting the benefit of having multiple networks through iterative pruning for constructing ensembles of networks.

\iffalse 
\paragraph{Other methods} Network quantization methods aim to improve the latency and energy consumption of networks by quantizing weights of those networks to low-precision fixed-point numbers. \cite{rastegari2016xnor} propose low-precision weights networks namely BWN and XNOR-Net that achieve comparable results with their counterparts, which have full-precision weights, on large-scale datasets. Quantization benefits both conventional computional devices as well as specialized accelerators. Hence, many methods in this line of work have been deployed in practice e.g. Google TPU\cite{jouppi2017datacenter}, TensorRT \cite{migacz20178} and so on.
\fi 

%------------------------------------------------------------------------- 
% KNOWLEDGE DISTILLATION
\section{Knowledge Distillation}
Consider the classification problem in which we need to determine the correct category for input image $\mathbf{x}$ among $M$ classes. The probability of class $m$ for sample $\mathbf{x}_n$ given by neural network $f$ parameterized by $\boldsymbol\theta$ is computed as:
\begin{equation}
  p_m(\mathbf{x}_n; \boldsymbol{\theta},\tau) = \frac{\exp( \frac{f_m(\mathbf{x}_n;\boldsymbol{\theta})}{\tau} )}{\sum_{i=1}^M \exp(\frac{f_i(\mathbf{x}_n;\boldsymbol\theta)}{\tau})}
\end{equation}
Where $\tau$ is the temperature of softmax function, higher values of $\tau$ lead to softer output distribution. Conventional approaches optimize the parameters $\boldsymbol\theta$ by sampling mini-batches $\mathcal{B}$ from the dataset and update the parameters to minimize cross-entropy objective:
\begin{equation}
  \label{equation:supervised_loss}
  \mathcal{L}_{NCE}(\mathcal{B};\boldsymbol\theta) = -\frac{1}{N}\sum_{n=1}^N\sum_{m=1}^M y_m \log p_m(\mathbf{x}_n;\boldsymbol\theta, 1)
\end{equation}

The target distribution of a sample is usually represented by \textit{one-hot vector} i.e. only the true class is $1$ and all other classes are $0$. Since input images might differ in term of noise, complexity, and multi-modality, enforcing networks to excessively fit the delta distribution of ground truth for all samples might deteriorate their generalization. Besides that, the similarity between classes provides rich information for learning and potentially prevent overfitting \cite{yang2019training}. Knowledge distillation \cite{bucilua2006model, hinton2015distilling} uses a trained (\textit{teacher}) network, which usually has high capacity, to guide the training of other (\textit{student}) network.  Let $q_m(\mathbf{x}_n)$ be the probability of class $m$ for image $\mathbf{x}_n$ given by the teacher network, which is parameterized by $\boldsymbol\psi$. The objective function of knowledge distillation is defined as:
\begin{equation}
  \mathcal{L}_{KD}(\mathcal{B};\boldsymbol\theta, \tau, \boldsymbol\psi) = -\frac{\tau^2}{N}\sum_{n=1}^N\sum_{m=1}^M q_m(\mathbf{x}_n;\boldsymbol\psi, \tau) \log \frac{q_m(\mathbf{x}_n;\boldsymbol\psi, \tau) }{p_m(\mathbf{x}_n;\boldsymbol\theta, \tau)}
\end{equation}
In case the teacher is an ensemble of $K$ networks, the target distribution of knowledge distillation is the average of outputs of all networks: $ \label{equation:avg_prob} \bar{q}_m(\mathbf{x}_n;\boldsymbol\psi_{1:K}, \tau) = \frac{1}{K}\sum_{k=1}^K q_m(\mathbf{x}_n; \boldsymbol\psi_k,\tau)$.

An alternative approach is optimizing the mean of Kullback-Leibler divergence between the student and each teacher network:
\begin{equation}
  \label{equation:kd_loss}
  \mathcal{L'}_{KD}(\mathcal{B};\boldsymbol\theta, \tau, \boldsymbol\psi_{1:K}) = -\frac{\tau^2}{KN}\sum_{n=1}^N\sum_{m=1}^M\sum_{k=1}^K q_m(\mathbf{x}_n;\boldsymbol\psi_k, \tau) \log \frac{q_m(\mathbf{x}_n;\boldsymbol\psi_k, \tau) }{p_m(\mathbf{x}_n;\boldsymbol\theta, \tau)}
\end{equation}
We experimented with two above objectives but did not observe significant difference in performance of student networks, thus, we only report results of the \textit{second} approach.

%------------------------------------------------------------------------- 
% SNAPSHOT
\section{Snapshots of Iterative Pruning}
%\paragraph{Essential of Iterative Pruning} Layers or weights in networks could be removed according to several criterions: $\ell_1$-magnitude of weights \cite{li2016pruning}, entropy of activations, accurate reduction, geometric median \cite{he2019filter}, batchnorm \cite{ye2018rethinking}, Taylor expansion \cite{molchanov2016pruning},... However, these criterions are rough estimation of weight importance, thus, having a lot of noise. Specifically, pruning algorithms rely on Taylor expansion would break down when the number of pruned filters increase greatly \cite{wang2019eigendamage}. In order to cope with this issue, instead of pruning a large number of weights at once, we prune the network with small number of weights and retrain iteratvely until reaching the desired compression ratio. 

\begin{figure}
  \centering\small
  \includegraphics[width=0.8\linewidth]{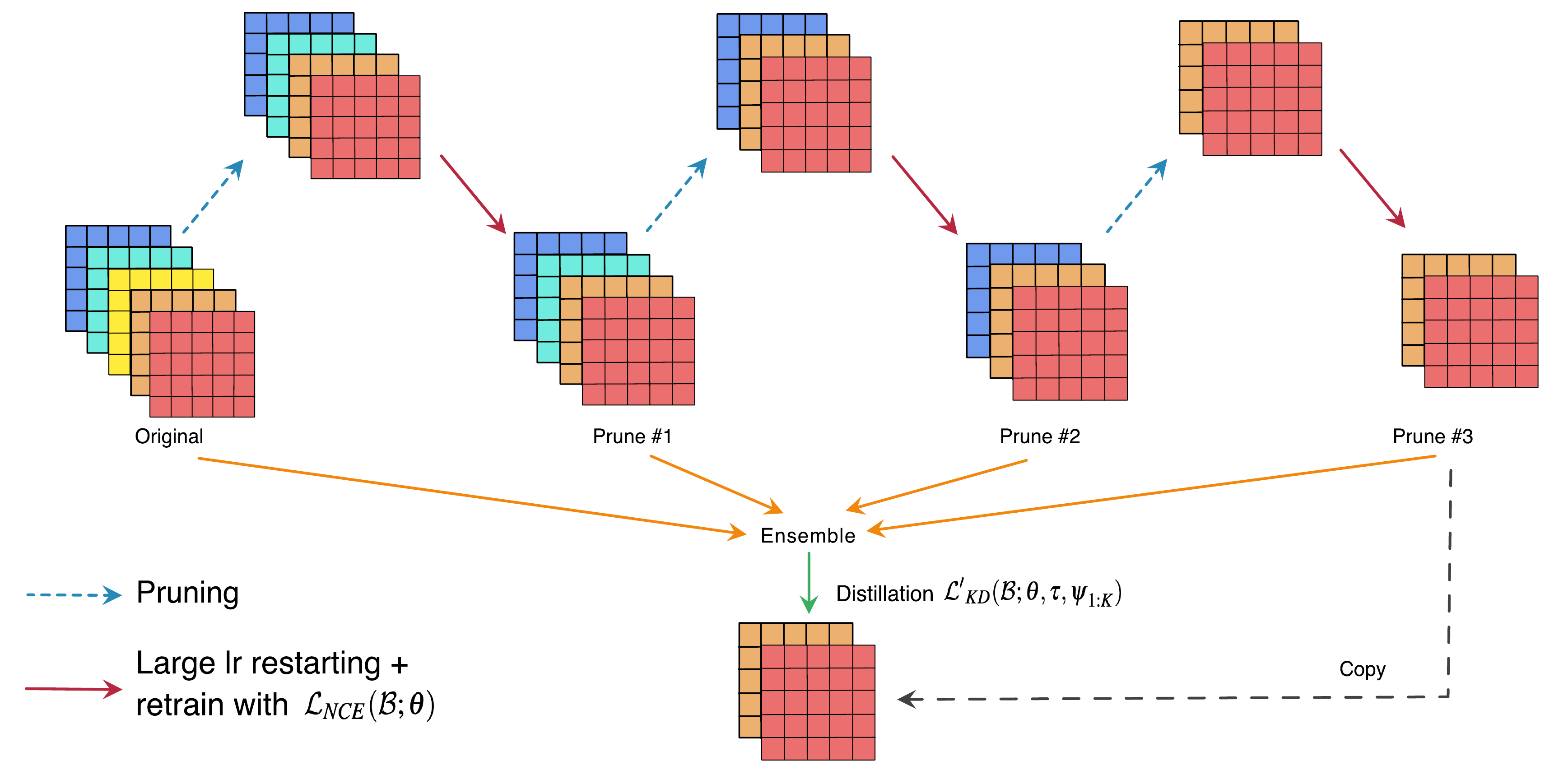}
  \caption{Overview of our approach combining the advantage of knowledge distillation, ensembles of networks, and network pruning. At the start, we prune the filters/weights according to some criteria ($\ell_1$-norm, Taylor approximation,...). With KESI, we retrain the pruned networks with large learning rate and minimize the conventional supervised loss function. Once we achieve the desired pruning ratio, we use knowledge distillation to transfer the knowledge from ensembles of snapshots of iterative pruning to the final model.}
  \label{fig:pipeline}
\end{figure}
In contrast to previous works, which mainly focus on the aforementioned usage of iterative pruning (i.e. alleviating the noise of weight's importance estimation), we exploit the benefits of generating multiple models varying in structure and capacity to construct strong ensembles.

Inspired by the prior works of \cite{smith2015no,loshchilov2016sgdr} in which the authors show that promising local optimums could be found in a small number of epochs after restarting the learning rate. Furthermore, Huang \etal \cite{huang2017snapshot} demonstrate that utilizing large learning rate restarting during training can construct strong ensembles without much additional cost. 

Broadly speaking, the performance of ensembles depends on: the performance of individual network and the diversity of them. On the other hand, network pruning generates snapshots varying in structure and achieving competitive performance. \textit{Hence, if pruned networks could achieve minimal loss in predictive power relative to the original network, the ensemble of them could potentially outperforms the ensemble of networks having identical architecture (and trained with large learning rate restarting)}.

Prior works such as \cite{han2015learning, liu2018rethinking, molchanov2016pruning} retrain the pruned networks for $T$ more epochs with a fixed learning rate, which is usually the final learning rate of the training. However, this approach might result in multiple snapshots being stuck in similar local optimums, thus, leading to very weak ensembles as shown in our experiments. Similar to \cite{huang2017snapshot}, we adopt the large learning rate restarting at each pruning iteration to encourage each snapshot to converge to different optimum. For learning rate restarting, we utilize the One-cycle policy \cite{smith2019super}, which is proved to increase convergence speed of several models. Due to the similarity of our proposed method and \textit{Snapshot Ensembling}~\cite{huang2017snapshot}, we refer to each pruning and retraining step as a \textit{cycle}. One-cycle policy adjusts learning rate at each mini-batch update and has two phases:

\textsc{Increasing learning rate} The learning rate and momentum of optimizer will be initialized to $\eta_{initial}$ and $\beta_{initial}$ respectively. During the first $T$ iterations of fine-tuning, learning rate and momentum gradually increase from initial values to $\eta_{max}$, $\beta_{max}$. The learning rate and momentum at $i$-th step with \textit{cosine annealing strategy} are given by:
\begin{equation}
      \eta_i = \eta_{max}+\frac{\eta_{initial}-\eta_{max}}{2}(1+\cos(\frac{i}{T}\cdot\pi))
\end{equation}

\begin{equation}
  \beta_i = \beta_{max}+\frac{\beta_{initial}-\beta_{max}}{2}(1+\cos(\frac{i}{T}\cdot\pi))
\end{equation}

\textsc{Decreasing learning rate}
After $T$ iterations, learning rate and momentum will be gradually decreased from $\eta_{max}$ and $\beta_{max}$ to $\eta_{min}$ and $\beta_{min}$ in $L-T$ iterations where $L$ is total number of iterations for fine-tuning.
\begin{equation}
  \eta_i = \eta_{min}+\frac{\eta_{max}-\eta_{min}}{2}(1+\cos(\frac{i-T}{L-T}\cdot\pi))
\end{equation}

\begin{equation}
  \beta_i = \beta_{initial}+\frac{\beta_{max}-\beta_{initial}}{2}(1+\cos(\frac{i-T}{L-T}\cdot\pi))
\end{equation}
It is worth noticing that differs from previous works \cite{huang2017snapshot, yang2019snapshot}, which use \textit{cosine annealing schedule}, by using One-cycle policy, we also \textit{"warm-up"}  learning rate at the start of each cycle. In our experiments, warming up learning rate is extremely important to achieve high accuracy with deep and large networks.

Surprisingly, retraining with One-cycle policy does not only generate significantly stronger ensembles, but also consistently \textbf{outperforms} the standard strategy for finetuning in terms of predictive accuracy of individual snapshots. We hypothesize that the (local) optimums of pruned networks are actually far from those of original networks, thus, large learning rate is needed to guarantee the convergence of pruned networks. We leave rigorous evaluation to investigate this phenomenon for future works.

%------------------------------------------------------------------------- 
% PIPELINE
\section{Effective Pipeline for Model Compression}
Since we already obtain strong ensembles during pruning, it is straightforward to distill the knowledge from them to the final pruned network. Our proposed pipeline can be summarized as follow:

\begin{algorithm}
    \begin{enumerate}
        \item \textsc{Train} the baseline model to completion. \\
        \item \textsc{Prune} redundant weights of the network based on some criteria.\\
        \item \textsc{Retrain} the pruned network with \textbf{large learning rate}. \\
        \item \textsc{Repeat} step 2 and 3 until desired compression ratio is reached. \\
        \item \textsc{Distill} knowledge from ensembles of snapshots of pruning. \\
    \end{enumerate}
\caption{Knowledge Distillation from Ensemble of Snapshots of Iterative pruning}
\end{algorithm}

From now, we refer to our pipeline for model compression as \textit{Knowledge Distillation from Ensembles of Snapshots of Iterative Pruning} (KESI). An overview of our approach is depicted in Figure \ref{fig:pipeline}. Our approach is extremely simple, easy to implement and can be adopted with any pruning mechanisms. We discuss the reasons why ensembles of snapshots of pruning are naturally suited for knowledge distillation. 

\textbf{Quality of Teacher} In knowledge distillation, student can either learn to jointly optimize the supervised loss (Equation \ref{equation:supervised_loss}) and knowledge distillation loss (Equation \ref{equation:kd_loss}) or only optimize the distillation objective. In the former case, if the teacher is poorly trained, mathematically speaking, the two objectives will conflict with each other. In the latter case, a poor teacher provides weak supervision (noisy label), making it's harder to learn from the student's perspective. Furthermore, ensembles provide more robust predictions on noisy labeled datasets \cite{lee2019robust} and out-of-distribution examples  \cite{lakshminarayanan2017simple}.

\textbf{Student and Teacher Gap} Although ensembles of snapshots have superior performance than the original network, it is not sufficient to guarantee the improvement in the performance of the student network with Knowledge Distillation. In fact, many works such as \cite{mirzadehimproved, cho2019efficacy, yang2019training} show that a powerful teacher might impair the performance of its student if there is a large gap between their predictive powers. However, ensembles of snapshots of pruning consist of models varying in capacity. Hence, teacher's predictions of hard-to-learn samples (because of their complexity, multi-modality) will have softer distributions as the small networks could not "remember" those samples and would be more uncertain about them.

In this work, we only investigate knowledge distillation from ensembles of fixed-weights teachers, however, we can also jointly train all models and allow them to guide each other, which is referred to as \textit{deep mutual learning} \cite{zhang2018deep}. 

\section{Experiments}
We conduct experiments on CIFAR-10, CIFAR-100 \cite{krizhevsky2009learning} and Tiny-Imagenet \footnote{\url{https://tiny- imagenet.herokuapp.com}} datasets. 

The two CIFAR datasets \cite{krizhevsky2009learning} consist of colored natural images sized at $32\times32$ pixels. CIFAR-10 (C10) and CIFAR-100 (C100) images are drawn from 10 and 100 classes, respectively. For each dataset, there are 50,000 training images and 10,000 images reserved for testing.

The Tiny ImageNet dataset consists of a subset of ImageNet images \cite{deng2009imagenet}. There are 200 classes, each of which has 500 training images and 50 validation images. Each image is resized to $64 \times 64$ and augmented with random crops, horizontal mirroring, and RGB intensity scaling.

We run each experiment 3 times then report mean and standard deviation of each network. In our experiments, we prune all networks in 5 \textit{cycles} unless otherwise stated.
\subsection{Experiment setup}
\subsubsection*{Training baselines}
We adopt the training and pruning code from \cite{liu2018rethinking} \footnote{\url{https://github.com/Eric-mingjie/rethinking-network-pruning}}. We train all networks with Stochastic Gradient Descent (SGD), learning rate is dropped from $0.1$ to $0.01$ at $50\%$ training and to $0.001$ at $75\%$. The batch size is set to $128$ and weight decay is $0.0001$ similar to \cite{he2016deep, he2016identity}.

\textbf{CIFAR} In order to create strong baseline models, we extend the training schedule of all models to 300 epochs. For WideResnet, we use same configurations as described in \cite{zagoruyko2016wide}.

\textbf{Tiny-Imagenet} we adopt Pytorch’s pretrained models on ImageNet and only replace the last fully-connected layer and train networks for $T = 100$ more epochs. We warm up learning rate from $0.01$ to $0.1$ in $10$ epochs. Other configurations are adopted from CIFAR training recipe.
\subsubsection*{Pruning}
\textbf{Structured pruning} we use $\ell_1$-norm based filters pruning \cite{li2016pruning} for simplicity. In each layer, a fixed number of filters having smallest $\ell_1$-norm will be pruned. Since the bulk of networks tend to be last layers, we increase the percentage of filters that will be pruned as the layer goes deeper to achieve higher compression ratio. \\
\textbf{Unstructured pruning}, we exploit (global) magnitude-based weight pruning \cite{han2015learning} i.e. pooling parameters across all layers and pruning weights with lowest magnitude. Specifically, we only prune parameters of convolutional layers similar to \citep{liu2018rethinking}.
\subsubsection*{Retraining}
The budget for fine-tuning of each cycle is $T=40$ and $T=25$ epochs on CIFAR and Tiny-Imagenet datasets respectively regardless of model architectures. In \textit{standard policy}, the learning rate is set to 0.001 and fixed during retraining. 

For \textit{One-cycle policy}, we set the initial learning rate $\eta_{initial} = 0.01$, gradually increase it to the maximum learning rate $\eta_{max} = 0.1$ in $10\%$ of total (retrain) epochs, then decrease it to the minimum learning rate $\eta_{min} = 0.0001$ for remaining epochs. Other configurations are identical to those of training.

\subsubsection*{Knowledge Distillation}
We use \textit{Adam} optimizer \cite{kingma2014adam} for ensemble distillation since it gives better results than vanilla SGD in our experiments. For knowledge distillation, we also adopt One-cycle policy where we set $\eta_{initial}, \eta_{max}, \eta_{min}$ to $1\mathrm{e}{-4}, 1\mathrm{e}-3, 1\mathrm{e}{-}6$ respectively. We do not explicitly use regularization for knowledge distillation. Other configurations \eg batch size, number of retraining epochs,... are similar to normal finetuning. 

In our experiments, we use temperature $\tau=5$. The teachers i.e. ensembles of snapshots consist of 6 models including the original (unpruned) network and 5 snapshots of pruning.
\subsection{Results}
\subsubsection{Effectiveness of large learning rate}
%%%%%%%%%%%%%%%%%%%%%%%%%%%%%%%%%%%%%%%%%%%
% COMPARING 

\begin{figure}[t]
  \centering
  \resizebox{\linewidth}{!}{
    \begin{tabular}{ccc}
      \includegraphics[width=0.45\linewidth]{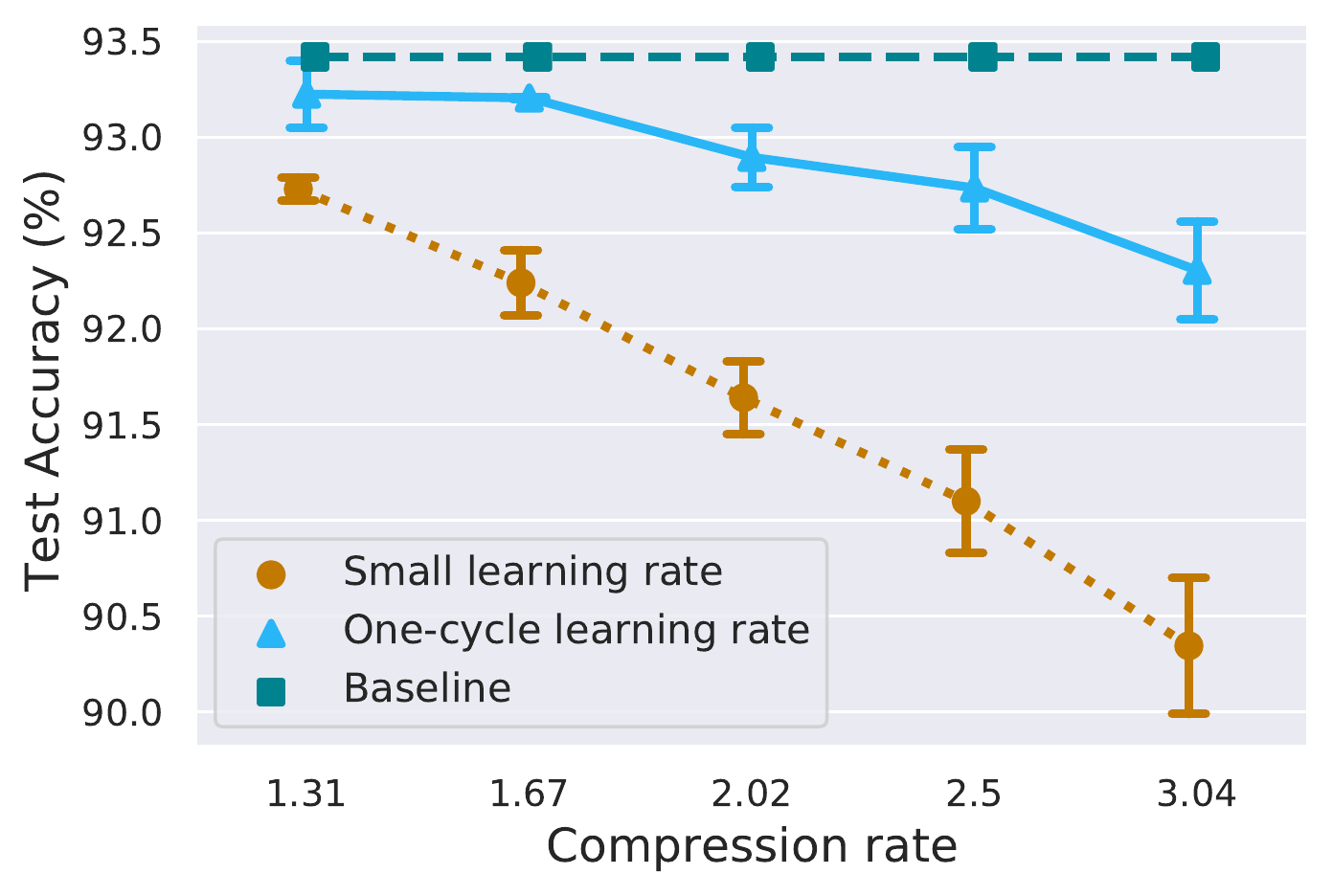} &
      \includegraphics[width=0.45\linewidth]{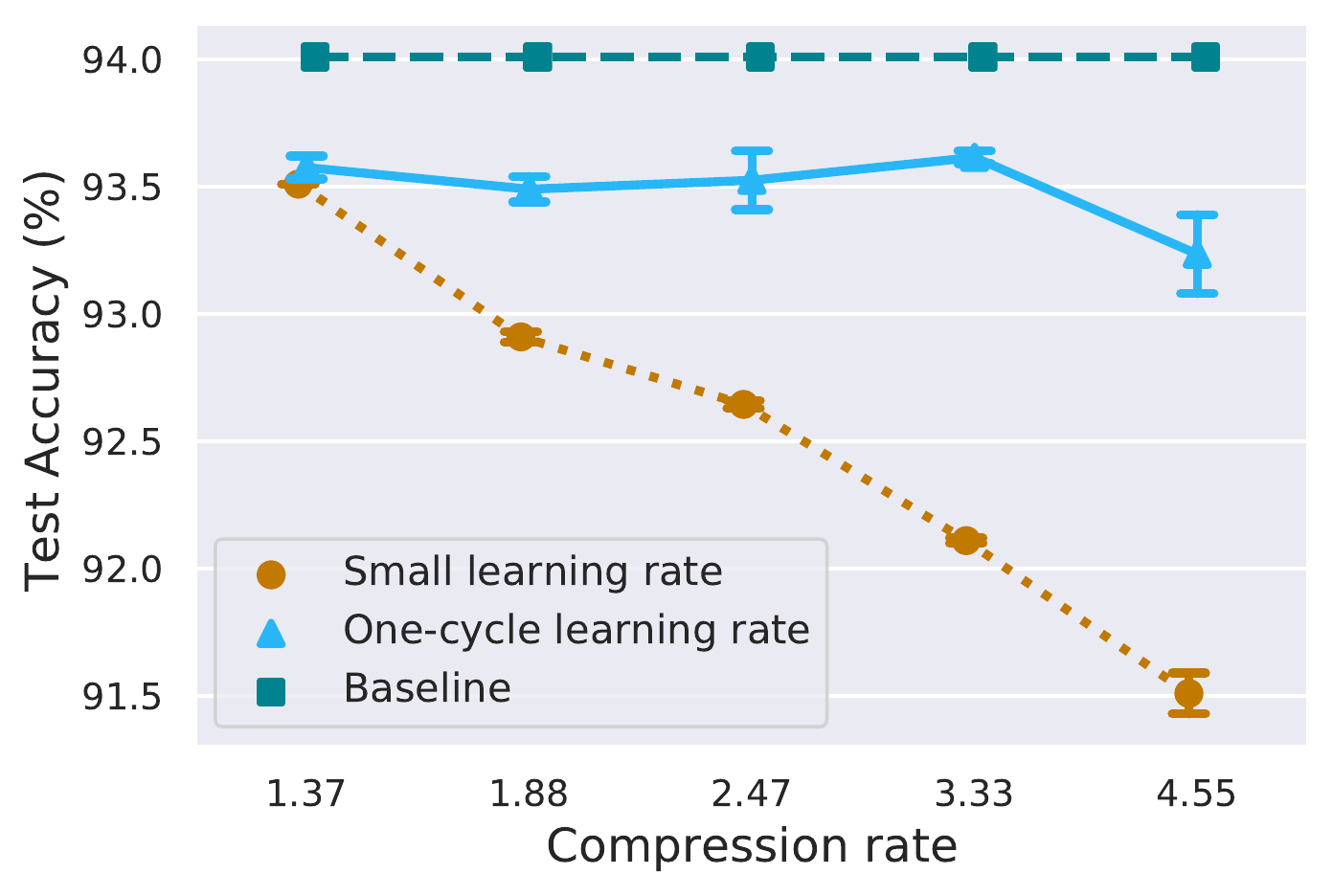} &
      \includegraphics[width=0.45\linewidth]{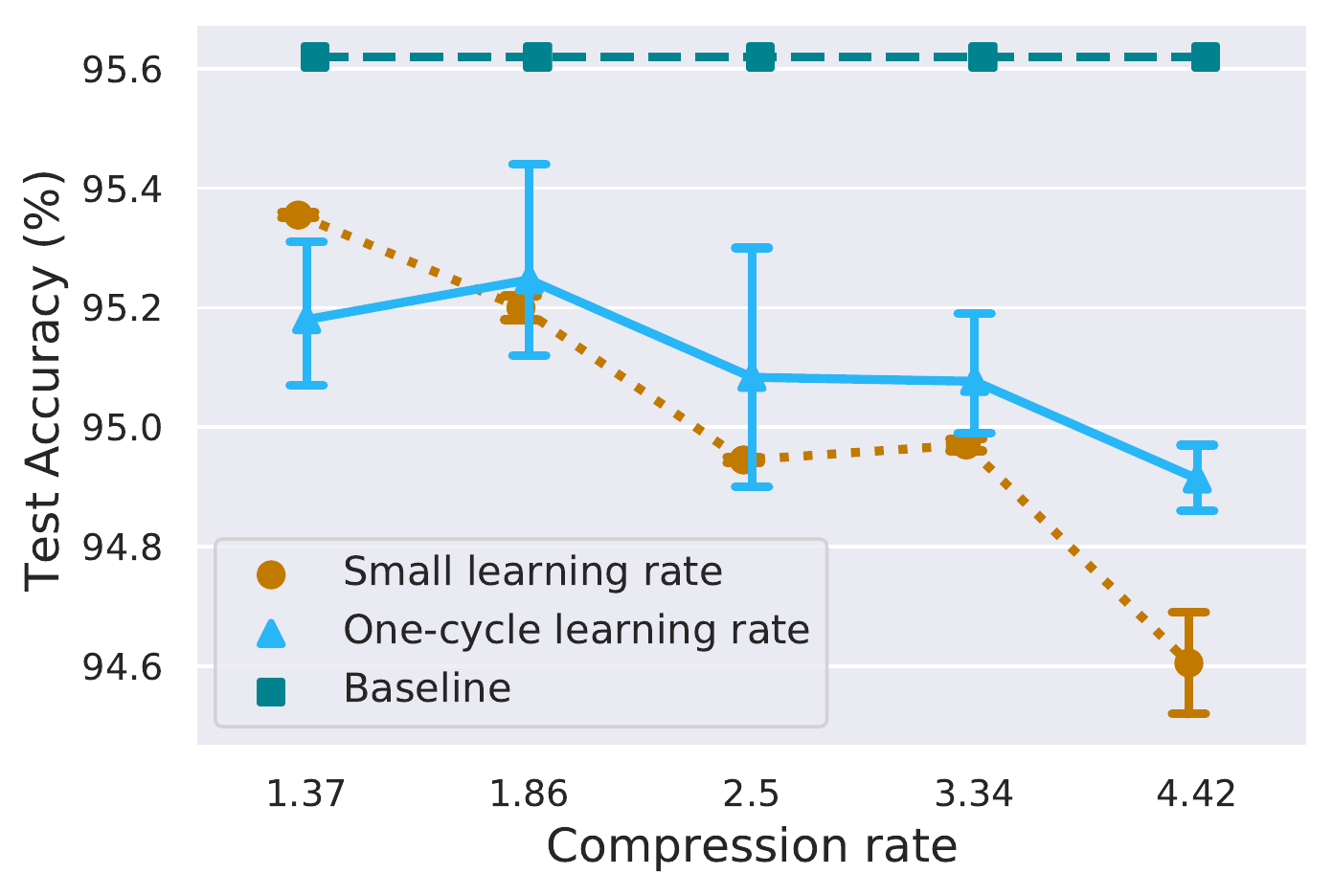} \\
      (a) Resnet-56 on CIFAR-10 & (b) Resnet-110 on CIFAR-10 & (c) WideResnet-16-8 on CIFAR-10 \\
    \end{tabular}
  }
  \resizebox{\linewidth}{!}{
    \begin{tabular}{ccc}
      \includegraphics[width=0.45\linewidth]{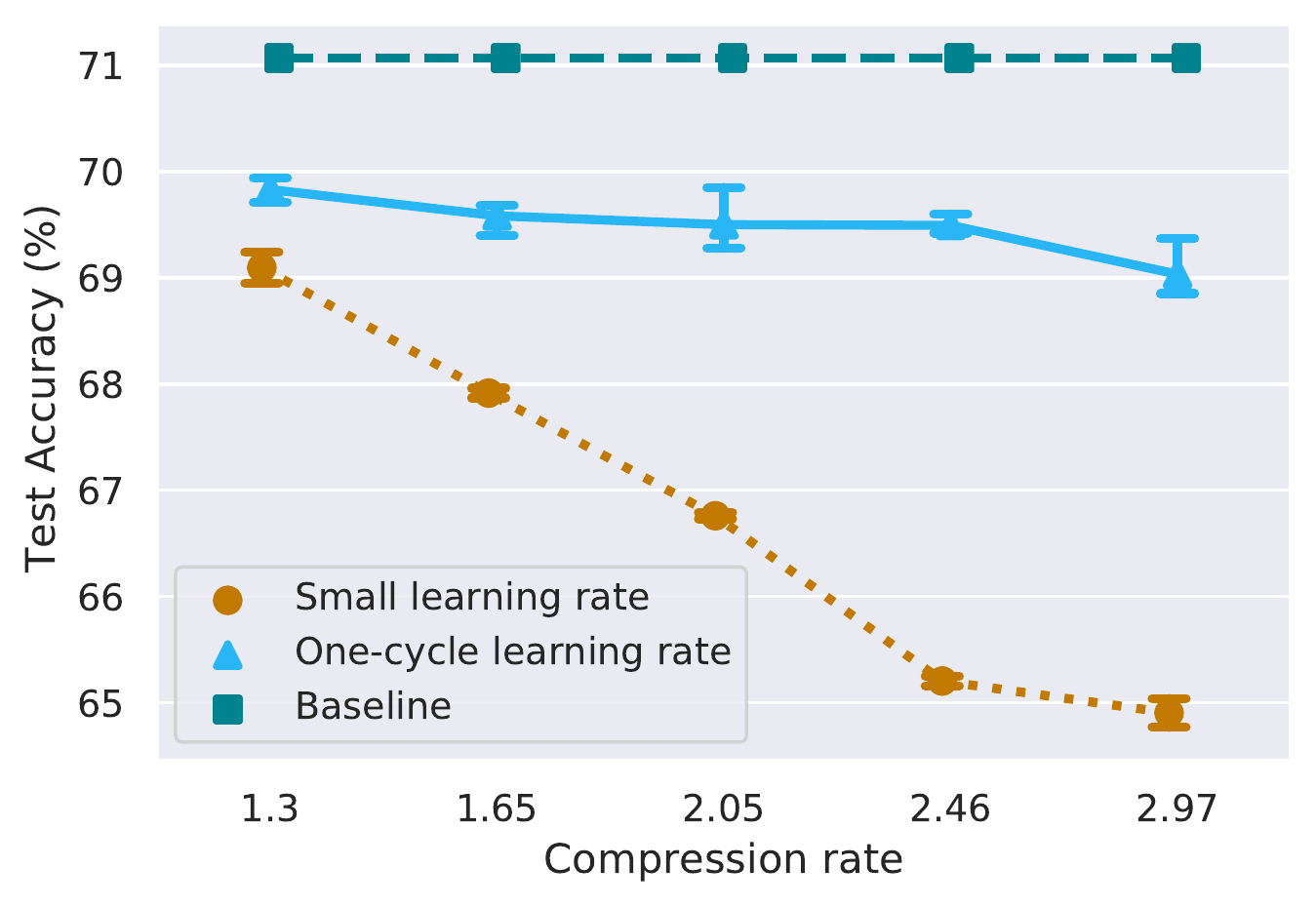} &
      \includegraphics[width=0.45\linewidth]{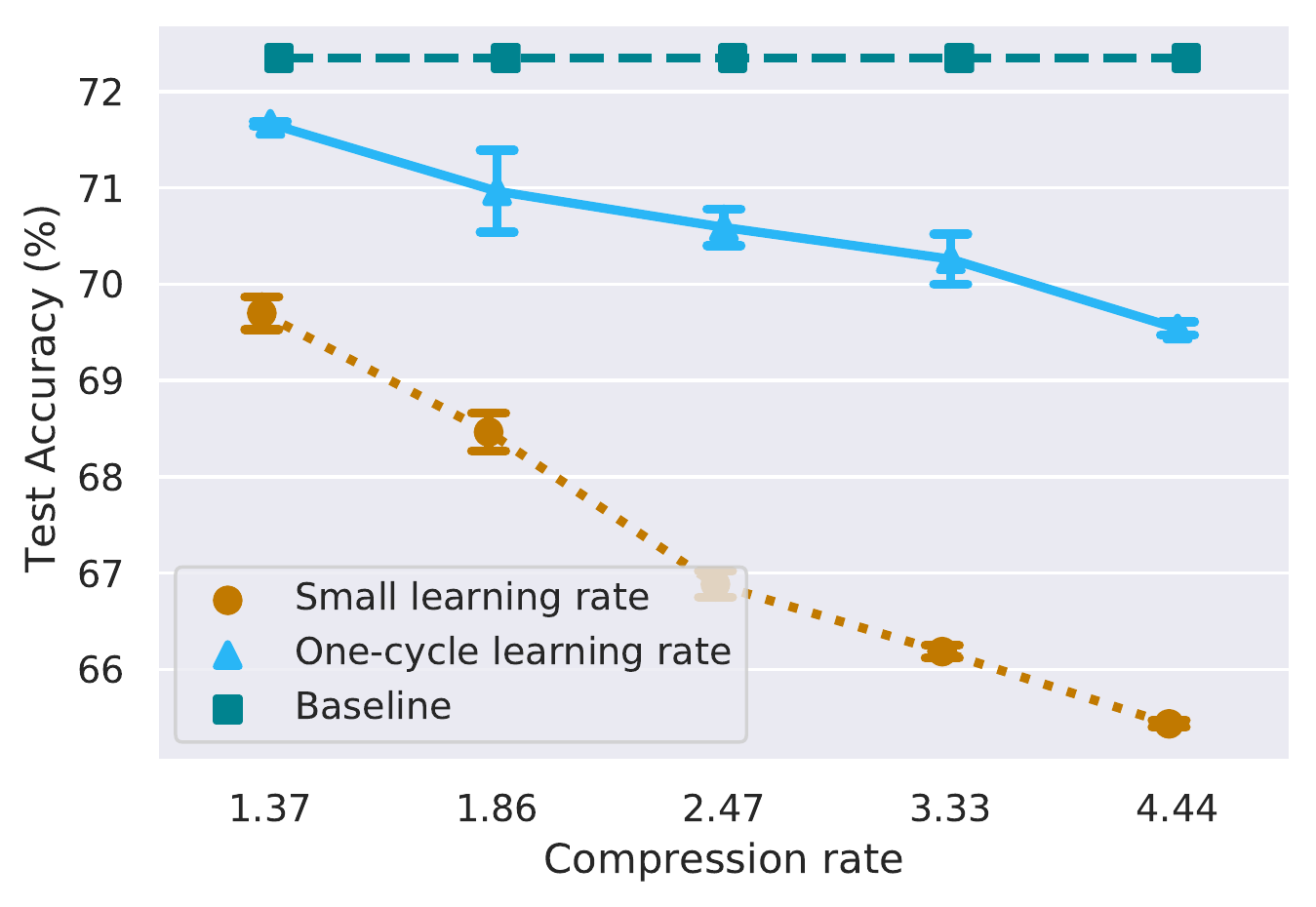} &
      \includegraphics[width=0.45\linewidth]{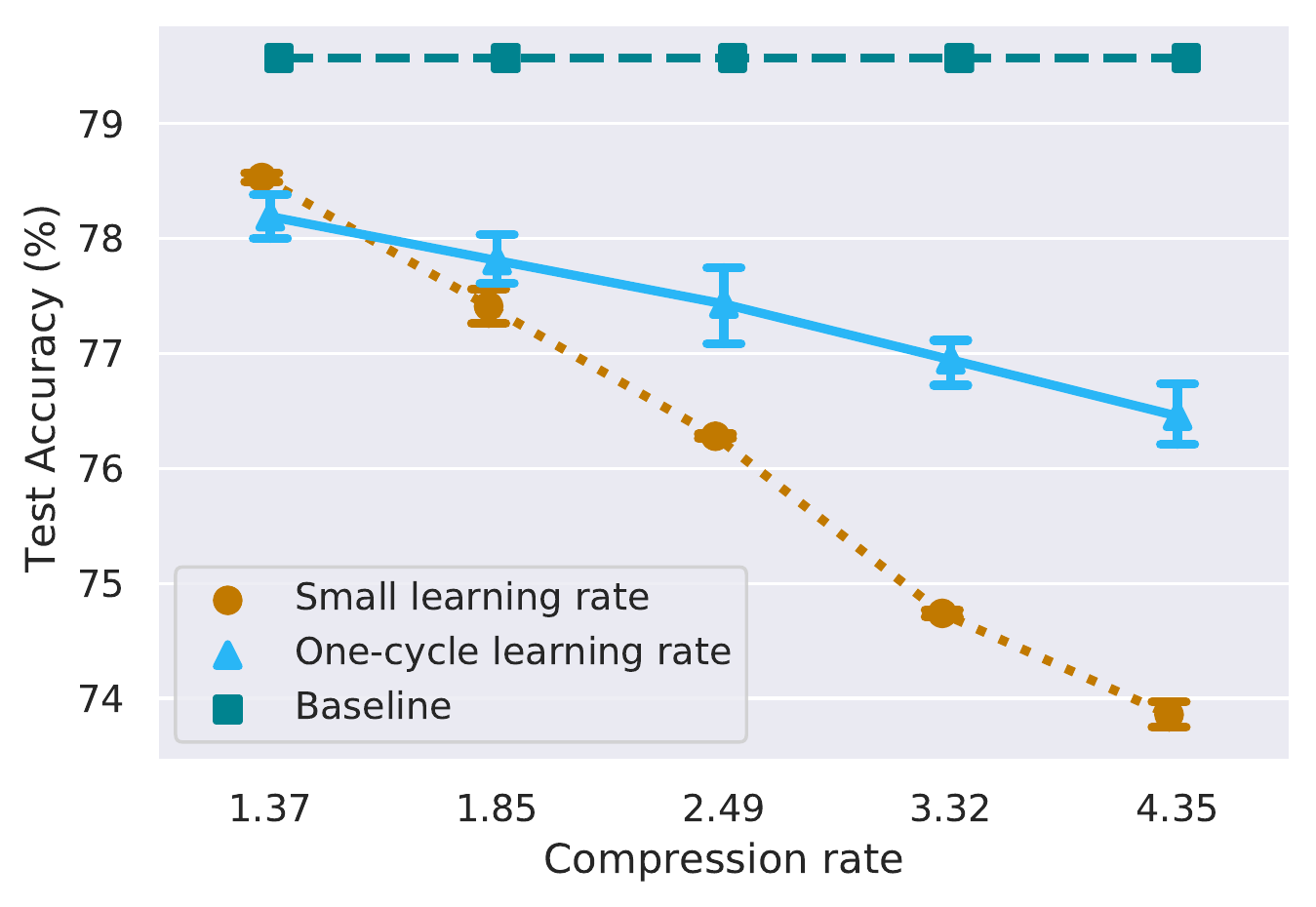} \\
      (a) Resnet-56 on CIFAR-100 & (b) Resnet-110 on CIFAR-100 & (c) WideResnet-16-8 on CIFAR-100 \\
    \end{tabular}
  }
  \caption{\textbf{$\ell_1$-norm filters pruning}~\cite{li2016pruning} with standard small, fixed learning rate and One-cycle learning rate.}
  \label{fig:compare_one-cycle_standard_structured}
\end{figure}
\begin{figure}[t]
  \centering
  \resizebox{\linewidth}{!}{
    \begin{tabular}{ccc}
      \includegraphics[width=0.45\linewidth]{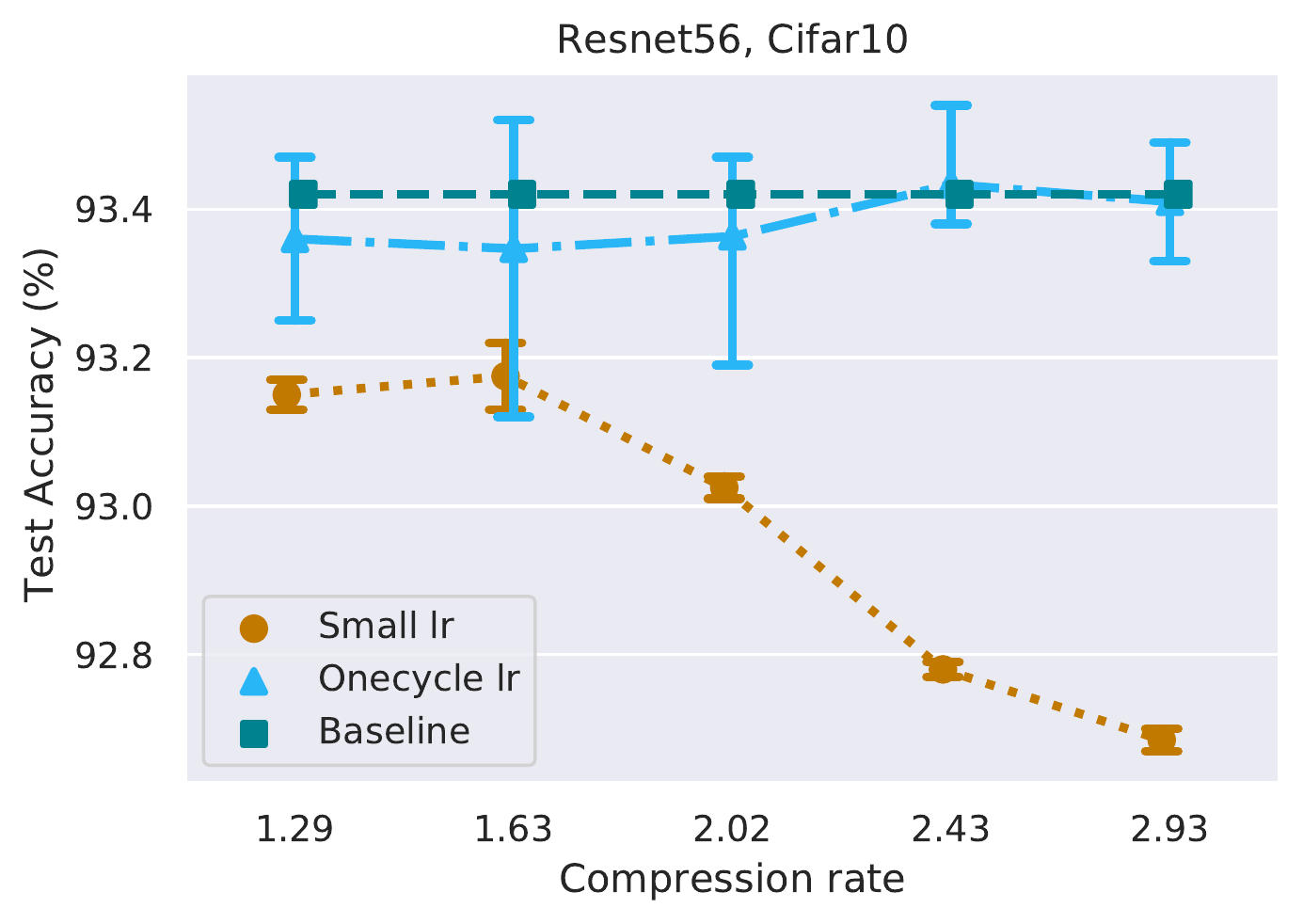} &
      \includegraphics[width=0.45\linewidth]{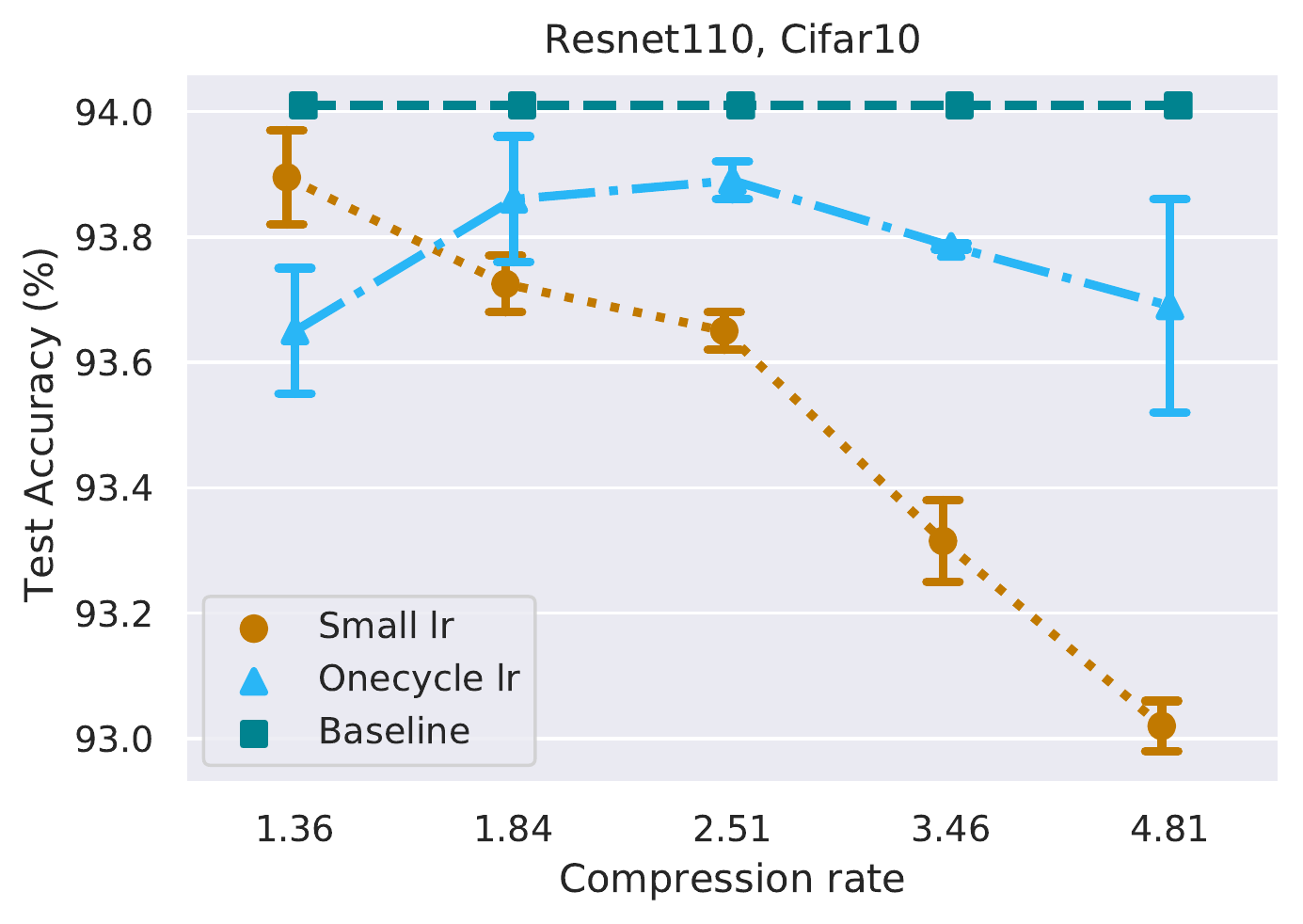} &
      \includegraphics[width=0.45\linewidth]{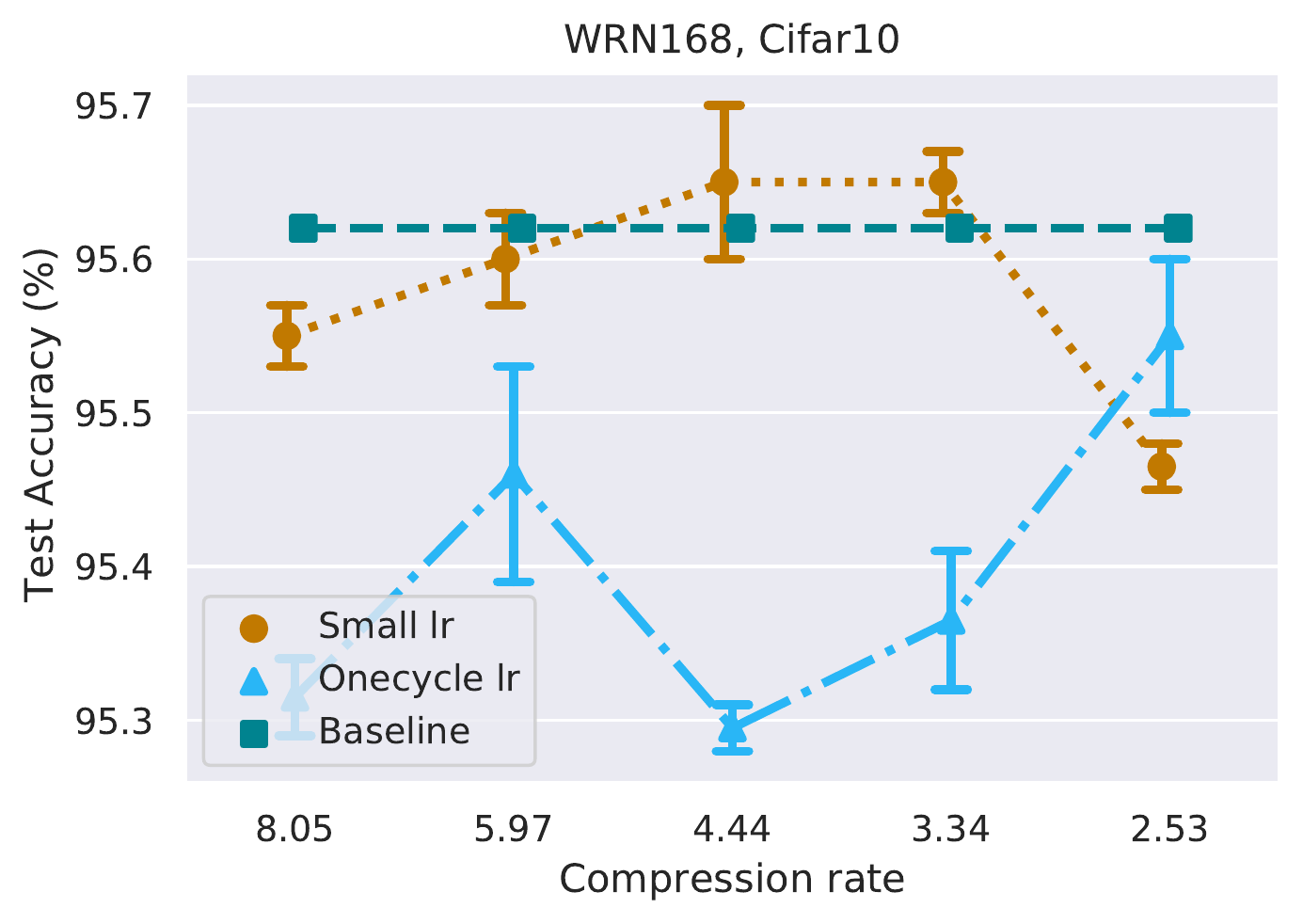} \\
      (a) Resnet-56 on CIFAR-10 & (b) Resnet-110 on CIFAR-10 & (c) WideResnet-16-8 on CIFAR-10 \\
    \end{tabular}
  }
  \caption{\textbf{Unstructured magnitude-based pruning}~\cite{han2015learning} with standard small, fixed learning rate and One-cycle learning rate.}
  \label{fig:compare_one-cycle_standard_unstructured}
\end{figure}

We conduct experiments to empirically evaluate the performance of pruned networks trained with large learning rate compare to networks fine-tuned with small learning rate. Figure \ref{fig:compare_one-cycle_standard_structured} and \ref{fig:compare_one-cycle_standard_unstructured} demonstrate results of pruned networks with different compression ratios for both structured and unstructured pruning. Exhaustive results are reported in supplementary documents. 

\subsubsection{Performance of ensembles of snapshots}
\begin{figure}[!t]
  \centering
  \includegraphics[width=\linewidth]{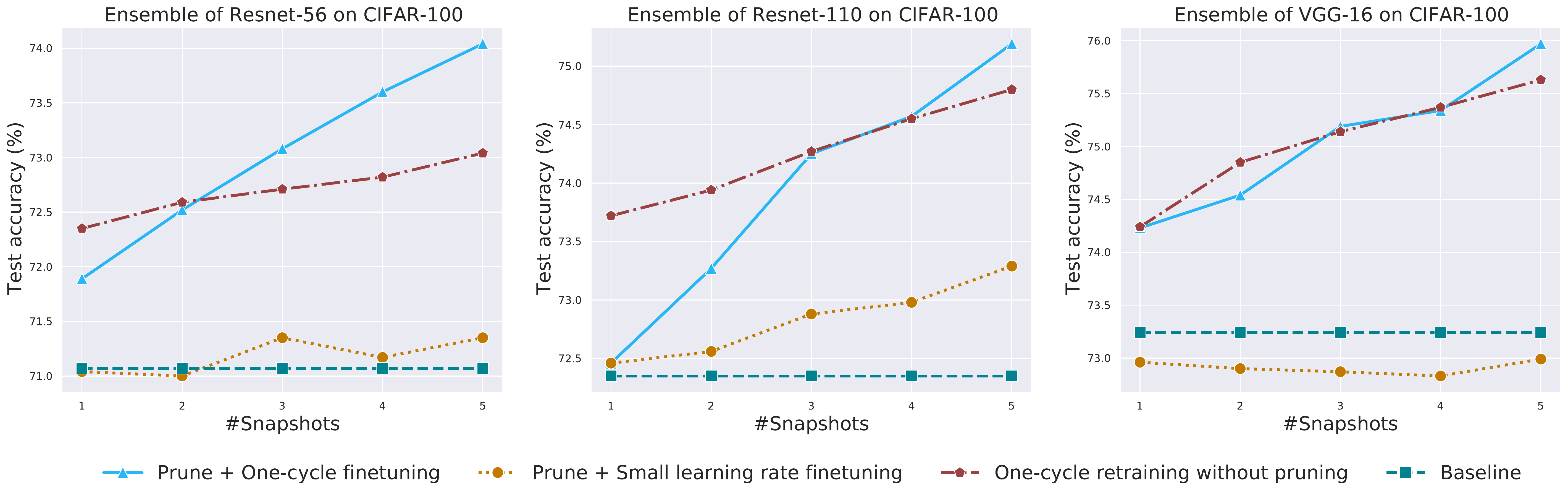}
  \caption{Performance of ensembles of snapshots with different approaches on CIFAR-100.}
  \label{fig:compare_ensemble}
\end{figure}

We compare the performance of ensembles of snapshots with different approaches: snapshots of pruned networks trained with small learning rate, snapshots of pruned networks trained with large learning rate restarting and snapshots of unpruned networks retrained with large learning rate (i.e. all snapshots have same architecture as the original network). Figure \ref{fig:compare_ensemble} presents the result of this experiment. 

We can see that although the network capacity is decreased at each \textit{cycle}, the ensembles of snapshots of iterative pruning achieve competitive or even better than snapshots of networks with same architecture. Detailed results of performance of ensembles are reported in supplementary documents.

\subsubsection{Performance of compact networks trained with our pipeline}

\begin{table}[t]
    \centering\small
    \resizebox{0.5\linewidth}{!}{
      \begin{tabular}{|l|l|cc|cc|}
        \hline
        Model                     & Methods                        & \% Params $\downarrow$ & \% FLOPs $\downarrow$ & baseline       & pruned                \\
        \hline\hline
        \multirow{5}*{Resnet-56}  & CP \citep{he2017channel}       & -                      & 50.6                 & 92.80          & 91.80                 \\
                                  & FPEC \citep{li2016pruning}     & 14.1                   & 27.6                 & 93.04          & 93.06                 \\
                                  & NISP \citep{yu2018nisp}        & 42.4                   & 35.5                 & 93.26          & 93.01                 \\
                                  & GAL-0.8 \citep{lin2019towards} & 65.9          & 60.2                 & 93.26          & 91.58                 \\
                                  & GBN \citep{you2019gate} &       66.7          & \textbf{70.3}                 & 93.10          & 93.07                 \\
								 & HRank \citep{lin2020hrank} &       42.4         & 50.0                 & 93.26          & 93.17                 \\
                                  & PFEC+KESI (our)                     & \textbf{67.1}          & 61.5       & \textbf{93.42} & $\mathbf{93.34\pm0.05}$ \\
        \hline
        \multirow{3}*{Resnet-110} & PFEC \citep{li2016pruning}     & 32.6                   & 38.7                 & 93.53          & 93.30                 \\
                                  & GAL-0.5 \citep{lin2019towards} & 44.8                   & 48.5                 & 93.50          & 92.74                 \\
                                  & HRank \citep{lin2020hrank} &       68.7         & \textbf{68.6}                 & 93.52          & 92.65                 \\
                                  & PFEC+KESI (our)                     & \textbf{77.5}          & 65.4        & \textbf{94.01} & $\mathbf{94.01\pm0.22}$ \\
        \hline
      \end{tabular}
    }
  \caption{Comparing performance of pruned networks with other approaches on CIFAR-10 dataset.}
  \label{table:comparison}

\end{table}
%%%%%%%%%%%%%%%%%%%%%%%%%%%%%%%%%%%%%%%%%%%
% REPORT
\begin{table}[!t]
  \centering
  \resizebox{\linewidth}{!}{
    \begin{tabular}{|l|c|cc|c|c||c|c|c|c|}
      \hline
      \multirow{2}*{Model}  & \multicolumn{5}{c||}{Structured Pruning} & \multicolumn{4}{c|}{Unstructured Pruning}  \\
      \cline{2-10}
      & Method   & \# Params(M) & \% MACs $\downarrow$ & C10  & C100 & Method & \# Params(M) & C10 & C100   \\
      \hline
      \hline
      \multirow{4}*{Resnet-56}       & baseline & 0.85         & 0.00                    & 93.42          & 71.07       & baseline & 0.85 & 93.42 & 71.07 \\
                                     & PFEC~\cite{li2016pruning}   & 0.28         &  61.5                       & $90.35\pm0.36$  & $64.91\pm0.14$ 
                                     & MWP~\cite{han2015learning}       & 0.29  & $92.69\pm0.02$   & $69.53\pm0.07$  \\
                                     & PFEC+One-cycle       & 0.28         &  61.5                       &  {\color{blue}$92.31\pm0.26$}  &{\color{blue} $69.03\pm0.24$}
                                     & MWP+One-cycle       & 0.29  & ${\color{blue}93.41\pm0.08}$   & ${\color{blue}70.46\pm0.30}$    \\
                                     & PFEC+KESI(our)       & 0.28         &  61.5                       & ${\color{blue}\mathbf{93.34\pm0.05}}$   & ${\color{blue}\mathbf{70.95\pm0.11}}$ 
                                     & MWP+KESI(our)       & 0.29  &  {\color{blue}$\mathbf{93.90\pm0.10}$}  &{\color{blue} $\mathbf{72.27\pm0.09}$}\\
      \hline
      \multirow{4}*{Resnet-110}       & baseline & 1.73         & 0.00                    & 94.01         & 72.35       & baseline & 1.73 & 94.01 & 72.35 \\
                                     & PFEC~\cite{li2016pruning}   & 0.39         &  65.38                       & $91.51\pm0.08$ & $65.44\pm0.04$
                                     & MWP~\cite{han2015learning}       & 0.36  & $93.02\pm0.04$   & $68.90\pm0.08$    \\
                                     & PFEC+One-cycle       & 0.39         &  65.38                       & {\color{blue}$93.24\pm0.16$}   & {\color{blue}$69.54\pm0.07$}
                                     & MWP+One-cycle       & 0.36  &  {\color{blue}$93.69\pm0.17$}   & {\color{blue}$71.59\pm0.30$}  \\
                                     & PFEC+KESI(our)       & 0.39         &  65.38                       & ${\color{blue}\mathbf{94.01\pm0.22}}$   & ${\color{blue}\mathbf{72.12\pm0.11}}$
                                     & MWP+KESI(our)       & 0.36  & ${\color{blue}\mathbf{94.44\pm0.11}}$   & ${\color{blue}\mathbf{73.12\pm0.25}}$    \\
      \hline
      \multirow{4}*{Preresnet-164}       & baseline & 1.70         & 0.00                    & 95.06          & 76.35       & baseline & 1.70 & 95.06 & 76.35 \\
                                     & PFEC~\cite{li2016pruning}   & 0.31         &  69.23                       & $92.05\pm0.11$   & $69.20\pm0.04$ 
                                     & MWP~\cite{han2015learning}       &   & &  \\
                                     & PFEC+One-cycle       & 0.31         &  69.23                       & ${\color{blue}94.15\pm0.06}$   & ${\color{blue}73.99\pm0.06}$  
                                     & MWP+One-cycle       &  & &\\
                                     & PFEC+KESI(our)       & 0.31         &  69.23                       & ${\color{blue}\mathbf{94.30\pm0.53}}$   & ${\color{blue}\mathbf{75.84\pm0.32}}$
                                     & MWP+KESI(our)       &   & &  \\
      \hline
      \multirow{4}*{WideResnet-16-8}       & baseline & 10.96         & 0.00                   & 95.62          & 79.57       & baseline & 10.96 & 95.62 & 79.57 \\
                                     & PFEC~\cite{li2016pruning}   & 2.48         &  64.52                       & $94.61\pm0.09$   & $73.82\pm0.10$ 
                                     & MWP~\cite{han2015learning}       & 2.53  & $95.47\pm0.07$   & $77.92\pm0.16$    \\
                                     & PFEC+One-cycle       & 2.48         &  64.52                       & ${\color{blue}94.91\pm0.04}$  & ${\color{blue}76.46\pm0.27}$
                                     & MWP+One-cycle       & 2.53  & ${\color{blue}95.55\pm0.05}$  & ${\color{blue}78.82\pm0.11}$    \\
                                     & PFEC+KESI(our)       & 2.48         &  64.52                       & ${\color{blue}\mathbf{95.68\pm0.12}}$  & ${\color{blue}\mathbf{79.01\pm0.20}}$
                                     & MWP+KESI(our)       & 2.53  & ${\color{blue}\mathbf{95.97\pm0.05}}$ & ${\color{blue}\mathbf{80.08\pm0.06}}$\\
      \hline
      \multirow{4}*{VGG-16}       & baseline & 14.99         & 0.00                    & 94.23          & 73.24       & baseline & 14.99 & 94.23 & 73.24 \\
                                     & PFEC~\cite{li2016pruning}   & 2.71         &  45.16                       & $93.88\pm0.12$   & $68.37\pm0.09$ 
                                     & MWP~\cite{han2015learning}       & 1.02  & $93.47\pm0.22$ & $68.39\pm0.21$ \\
                                     & PFEC+One-cycle       & 2.71         &  45.16                       & {\color{blue}$94.10\pm0.09$}  & ${\color{blue}71.95\pm0.04}$ 
                                     & MWP+One-cycle       & 1.02  & ${\color{blue}93.53\pm0.10}$ & ${\color{blue}71.74\pm0.15}$ \\
                                     & PFEC+KESI(our)       & 2.71        &  45.16                       &  ${\color{blue}\mathbf{94.59\pm0.09}}$   & ${\color{blue}\mathbf{73.52\pm0.20}}$
                                     & MWP+KESI(our)       & 1.02  & ${\color{blue}\mathbf{94.01\pm0.06}}$ & ${\color{blue}\mathbf{73.91\pm0.09}}$\\
      \hline
    \end{tabular}
  }
    \caption{Accuracy (\%) of pruned networks on CIFAR-10 and CIFAR-100 datasets trained with different strategies. PFEC (or MWP) are models pruned with $\ell_1$-norm filters pruning~\cite{li2016pruning} (or magnitude-based weights pruning~\cite{han2015learning}) and fine-tuned with small learning rate. PFEC/MWP+One-cycle are pruned networks retrained with large learning rate restarting. PFEC/MWP+KESI are pruned networks retrained with our pipeline}
    \label{table:kd_cifar}

\end{table}
%Tiny-Imagenet
\begin{table}[!t]
  \centering\small
  \makebox[0pt][c]{\parbox{\textwidth}{%
      \begin{minipage}[b]{0.5\hsize}\centering
        \caption{Performance of compact models on Tiny-Imagenet}
        \resizebox{\linewidth}{!}{
          \begin{tabular}{|l|c|cc|c|}
            \hline
            Model                    & Method   & \#Params (M) & MACs(G) & Acc          \\
            \hline
            \hline
            \multirow{4}*{Resnet-18} & baseline & 11.01        & 1.82                    & 67.22        \\
            & FPEC \cite{li2016pruning}      & 2.71         & 0.83                       & $61.06\pm0.32$ \\
            & FPEC+One-cycle       & 2.71         & 0.83                       & ${\color{blue}64.70\pm0.33}$ \\
                                     & FPEC+KESI (our)       & 2.71         & 0.83                       & ${\color{blue}\mathbf{66.87\pm0.26}}$ \\
            \hline
            \multirow{4}*{Resnet-34} & baseline & 21.39        & 3.68                    & 68.81        \\
            & FPEC \cite{li2016pruning}      & 5.40         & 1.57                       & $64.93\pm0.15$ \\
            & FPEC+One-cycle      & 5.40         & 1.57                       & ${\color{blue}67.26\pm0.21}$ \\
                                     & FPEC+KESI (our)      & 5.40         & 1.57                       & ${\color{blue}\mathbf{70.02\pm0.43}}$ \\
            \hline
          \end{tabular}
        }
        \label{table:kd_tinyimagenet}
      \end{minipage}
      \hfill
      \begin{minipage}[b]{0.47\hsize}\centering
        \caption{Knowledge distillation with ensembles teacher and single model teacher}
        \resizebox{\linewidth}{!}{
          \begin{tabular}{|l|l|cc|c|}
            \hline
            Model                     & Method         & \#Params (M) & C10         & C100           \\
            \hline
            \hline
            \multirow{3}*{Resnet-56}  & baseline       & 0.85        & 93.42       & 71.07          \\
                                      & single teacher & 0.28         & $93.13\pm0.04$ & $70.29\pm0.14$    \\
                                      & ensemble teacher      & 0.28         & $\mathbf{93.34\pm0.05}$       & $\mathbf{72.27\pm0.09}$   \\
            \hline
            \multirow{3}*{Resnet-110} & baseline       & 1.73        & 94.01       & 72.35 \\
                                      & single teacher & 0.39         & $93.48\pm0.05$ & $71.50\pm0.11$    \\
                                      & ensemble teacher      & 0.39         & $\mathbf{94.01\pm0.22}$       & $\mathbf{73.12\pm0.25}$   \\
            \hline
            \multirow{3}*{WRN-16-8}   & baseline       & 19.96        & 95.62       & 79.57 \\
                                      & single teacher & 2.48         & $95.37\pm0.21$ & $78.71\pm0.24$    \\
                                      & ensemble teacher       & 2.48         & $\mathbf{95.68\pm0.12}$          & $\mathbf{79.01\pm0.20}$   \\
            \hline
          \end{tabular}
        }
        \label{table:ablation_kd}
      \end{minipage}
    }}
\end{table}

In this section, we demonstrate that the smaller models trained with our pipeline (KESI) achieve comparable or even better results than the original model. Each final model is iteratively pruned and retrained in 5 \textit{cycles} with different strategies. Table \ref{table:kd_cifar} and \ref{table:kd_tinyimagenet} present the performance of compact models on CIFAR-10, CIFAR-100 and Tiny-Imagenet. Specifically, we compare the iteratively-pruned-models retrained with small learning rate, large learning rate and our pipeline (i.e. large learning rate + knowledge distillation). Our pipeline consistently outperforms the standard strategy by a large margin for both structured and unstructured pruning.

Although our approach is general and can be applied to any (iterative) pruning mechanism, we also give a comparison of model trained with our pipeline and conventional approaches in table \ref{table:comparison}. We conduct experiment to compare performance of student networks trained with single teacher (i.e. original/unpruned networks) and ensembles teacher in table \ref{table:ablation_kd} for ablation study. We can see that compact models learn from ensembles outperform those learn from a single teacher by a large margin. 
\section{Conclusion}
We propose a simple pipeline by slightly modifying the standard approach to acquire the advantages of network ensembles, knowledge distillation and network pruning. Our experiments show that small and compact networks trained with our pipeline significantly outperform the standard approach and create very strong baselines for model compression. Specifically, our method reduces nearly $80\%$ of parameters and $70\%$ FLOPs of several models by structured pruning without incurring loss in performance.\\
\textbf{Acknowledgement}
The authors thank anonymous reviewers and area chairs for their useful feedback. We also want to express our appreciation to Ms. Le Thi Tham Quynh for her valuable aids in the final preparation of the paper.

\bibliography{egbib}
\newpage
\begin{appendix}
\section{Results}
\begin{table}[!h]
	\caption{Results of iterative \textbf{$\ell_1$-norm Filters Pruning} \cite{li2016pruning} on  CIFAR-10 and CIFAR-100 datasets. The \textit{SLR} column presents the result of pruned networks finetuned with small learning rate while \textit{LLR} column shows the results of same networks finetuned with large learning rate. }
		\label{table:compare_lr_cifar10}
		\centering
		\resizebox{\linewidth}{!}{
		\begin{tabular}{|l|rc | cc | cc|}
			% \toprule
			\hline
			Model      & \#Param     &   \% MACs(G)$\downarrow$ & C10-SLR & C10-LLR & C100-SLR & C100-LLR\\
			\hline
			\hline
			Resnet-110 (baseline) & 1.73M  & 0.00  & 94.01  & 94.01 & 72.35  & 72.35 \\
			Resnet-110 \#1  &  1.26M  &   23.08  & \textbf{$93.51\pm0.05$} &  $93.58\pm0.05$ & $69.70\pm0.17$ &  $71.67\pm0.03$ \\
			Resnet-110 \#2  &  0.93M  &   38.46  & $92.91\pm0.02$ &  \textbf{$93.49\pm0.05$}  & $68.47\pm0.20$  &  $70.97\pm0.43$  \\
			Resnet-110 \#3  &  0.70M  &   50.00  & $92.65\pm0.02$ &  \textbf{$93.53\pm0.12$}  & $66.89\pm0.14$ &  $70.59\pm0.19$  \\
			Resnet-110 \#4  &  0.52M  &   57.69  & $92.11\pm0.01$ &  \textbf{$93.62\pm0.03$}  & $66.19\pm0.07$ &  $70.26\pm0.26$  \\
			Resnet-110 \#5  &  0.39M  &   65.38  & $91.51\pm0.08$ &   \textbf{$93.24\pm0.16$} & $65.44\pm0.04$ &   $69.54\pm0.07$\\
			Resnet-110 Ensemble  &  -  &  -   & 93.82 &    \textbf{$94.01\pm0.22$} & 73.32 &    \textbf{$75.33\pm0.14$}\\
			\hline
			Resnet-56 (baseline) & 0.85M  & 0.00  & 93.42  & 93.42 & 71.07  & 71.07 \\
			Resnet-56 \#1  &  0.66M  & 23.07  & $92.73\pm0.06$ & \textbf{$93.23\pm0.18$} &  $69.10\pm0.15$ &  $69.83\pm0.09$\\
			Resnet-56 \#2  &  0.52M &  30.77  & $92.24\pm0.17$ &  \textbf{$93.21\pm0.08$} & $67.92\pm0.05$ &   $69.58\pm0.13$ \\
			Resnet-56 \#3  &  0.42M  & 46.15  & $91.64\pm0.19$ &  \textbf{$92.90\pm0.16$} & $66.76\pm0.03$ &  $69.50\pm0.25$ \\
			Resnet-56 \#4  &  0.35M & 53.85   & $91.10\pm0.27$ &  \textbf{$92.74\pm0.22$} &  $65.205\pm0.05$ & $69.49\pm0.08$ \\
			Resnet-56 \#5  &  0.29M  &  61.54 & $90.35\pm0.36$  &  \textbf{$92.31\pm0.26$} & $64.91\pm0.14$  & $69.03\pm0.24$ \\
			Resnet-56 Ensemble  &  -  &  -   & 93.25 &    \textbf{$94.29\pm0.02$} & 71.37 &  $74.23\pm0.21$ \\
			\hline
			VGG-16 (baseline) & 14.99M & 0.00 & 94.23 & 94.23 & 73.24 & 73.24 \\ 
			VGG-16 \#1 & 9.46M & 0.00 & \textbf{$94.13\pm0.09$} & $93.95\pm0.02$ & $71.39\pm0.06$ & \textbf{$72.38\pm0.11$}\\ 
		    VGG-16 \#2 & 6.27M & 0.00 &  \textbf{$94.09\pm0.13$}  & $93.90\pm0.03$& $70.48\pm0.07$  & \textbf{$72.10\pm0.1$} \\ 
		    VGG-16 \#3 & 4.43M & 0.00 &  $94.09\pm0.04$  & $93.93\pm0.04$ &  $69.73\pm0.06$  & $72.28\pm0.11$ \\ 
		    VGG-16 \#4 & 3.36M & 0.00 &  $94.03\pm0.13$  & $93.89\pm0.10$ & $69.09\pm0.05$  & $72.22\pm0.19$\\ 
		    VGG-16 \#5 & 2.71M & 0.00 &  $93.88\pm0.12$  & $94.10\pm0.09$ & $68.37\pm0.09$  & $71.95\pm0.04$ \\ 
		    VGG-16 Ensemble & - & - &  94.29  & $95.04\pm0.07$  & $72.86\pm0.02$  & $75.93\pm0.06$ \\ 
		    \hline
		    PreResnet-164 (baseline) & 1.7M  & 0.00  & 95.06  & 95.06 & 76.35 & 76.35\\
		    PreResnet-164 \#1  &  1.09M  & 26.92 & $94.43\pm0.06$  & $94.92\pm0.05$ & $74.65\pm0.04$  & $76.20\pm0.07$ \\
		    PreResnet-164 \#2  &  0.74M  &  46.15 & $93.71\pm0.07$  &  $94.74\pm0.14$  &  $73.17\pm0.03$ & $75.87\pm0.03$  \\
		    PreResnet-164 \#3  &  0.54M  &  57.69  & $93.50\pm0.01$ &  $94.66\pm0.19$ & $71.89\pm0.01$ & $75.15\pm0.16$ \\
		    PreResnet-164 \#4  &  0.4M  & 65.38  & $92.61\pm0.02$ & $94.69\pm0.17$ & $70.50\pm0.09$ & $74.03\pm0.68$ \\
		    PreResnet-164 \#5  &  0.31M  &  69.23  &  $92.06\pm0.11 $ & $94.15\pm0.06$ & $69.20\pm0.04$ & $73.99\pm0.06$ \\
		    PreResnet-164 Ensemble  &  -  &  -   &  - &  $95.60\pm0.04$ &  - & $79.19\pm0.07$\\
		    \hline
	    	%WideResnet-28-10 (baseline)  & 36.65 & 0.00 & -  & - & 81.86  & 81.86 \\
	    	%WideResnet-28-10 \#1  & 26.71 & 0.00 & -  & - & -   & 80.62 \\
	    	%WideResnet-28-10 \#2  & 19.32 & 0.00 & -  & - & -  & 80.28 \\
	    	%WideResnet-28-10 \#3  & 14.55 & 0.00 & -  & - & -  & 79.96 \\
	        %WideResnet-28-10 Ensemble  & -  & - & -  & - & - & 82.53 \\
	        WideResnet-16-8 (baseline)  & 11.01  & 0.00 & 95.62 & 95.62 & 79.57 & 79.57 \\
	        WideResnet-16-8 \#1  &  8.01 & 20.00 & $95.36\pm0.01$  & $95.18\pm0.1$ & $78.52\pm0.04$  & $78.19\pm0.19$ \\
	        WideResnet-16-8 \#2  &  5.89 & 35.48 & $95.20\pm0.02$ & $95.25\pm0.14$  &  $77.46\pm0.14$  & $77.81\pm0.2$\\
	        WideResnet-16-8 \#3  &  4.38 & 47.74 & $94.95\pm0.01$ & $95.08\pm0.16$ & $76.29\pm0.02$  & $77.43\pm0.36$ \\
	        WideResnet-16-8 \#4  &  3.28 & 57.42 & $94.97\pm0.01$  & $95.08\pm0.08$ & $74.73\pm0.03$  & $76.95\pm0.21$\\
	        WideResnet-16-8 \#5  &  2.48 & 64.52 & $94.61\pm0.09$  & $94.91\pm0.04$ & $73.82\pm0.1$  &  $76.46\pm0.27$\\
	        WideResnet-16-8 Ensemble  & -  & - & 95.63  & $95.79\pm0.08$ & $79.22\pm0.03$  &  $80.45\pm0.14$\\
	    	\hline
		\end{tabular}}
	\end{table}
	
	\begin{table}[!h]
		\caption{Results of iterative \textbf{Weights pruning} \cite{han2015learning} on  CIFAR-10 and CIFAR-100 datasets. The \textit{SLR} column presents the result of pruned networks finetuned with small learning rate while \textit{LLR} column shows the results of same networks finetuned with large learning rate. }
		\label{table:wp_cifar}
		\centering
			\resizebox{\linewidth}{!}{
			\begin{tabular}{|l|c | cc | cc|}
				\hline
				Model      & \#Active Params (M)  &   C10-SLR & C10-LLR & C100-SLR & C100-LLR\\
				\hline
				\hline
				Resnet-56 (baseline) & 0  & 93.42 & 93.42 & 71.07  & 71.07 \\
				Resnet-56 \#1  & 0.66   & $93.15\pm0.02$  & $93.36\pm0.11$ & $70.95\pm0.05$ & $70.40\pm0.12$  \\
				Resnet-56 \#2  & 0.52   & $93.18\pm0.05$  & $93.35\pm0.17$  &  $70.78\pm0.01$ & $70.58\pm0.19$  \\
				Resnet-56 \#3  & 0.42     & $93.03\pm0.02$ & $93.36\pm0.12$  & $70.31\pm0.06$  &  $70.51\pm0.13$ \\
				Resnet-56 \#4  & 0.35    & $92.78\pm0.01$  & $93.43\pm0.08$ & $69.91\pm0.01$ &  $70.75\pm0.04$ \\
				Resnet-56 \#5  & 0.29   & $92.69\pm0.02$ &  $93.41\pm0.08$  &  $69.53\pm0.07$  &  $70.46\pm0.3$ \\
				Resnet-56 Ensemble  &  - & 93.39 & $94.15\pm0.04$  & $71.18\pm0.08$ &  $72.69\pm0.02$ \\
				\hline
				Resnet-110 (baseline) & 0  & 94.01 & 94.01 & 72.35  & 72.35 \\
				Resnet-110 \#1  &  1.27  & $93.90\pm0.08$  & $93.65\pm0.1$  & $72.30\pm0.07$  & $72.11\pm0.24$   \\
				Resnet-110 \#2  &  0.94 & $93.73\pm0.05$ & $93.86\pm0.1$ & $71.75\pm0.02$ &  $72.09\pm0.06$ \\
				Resnet-110 \#3  &  0.69 & $93.65\pm0.03$ & $93.89\pm0.03$ & $70.96\pm0.04$ &  $72.38\pm0.25$ \\
				Resnet-110 \#4  &  0.50  & $93.32\pm0.07$ & $93.79\pm0.01$ & $70.69\pm0.02$ &  $72.09\pm0.05$\\
				Resnet-110 \#5  &  0.36 & $93.02\pm0.04$ & $93.69\pm0.17$ & $68.90\pm0.08$ &  $71.59\pm0.30$ \\
				Resnet-110 Ensemble  &  -   & $93.98\pm0.04$  & $94.56\pm0.04$  & $72.51\pm0.06$  &  $74.19\pm0.02$ \\
				\hline
				WideResnet-110 (baseline) & 11.01  & 95.62 & 95.62 & 79.57  & 79.57 \\
				WideResnet-110 \#1  & 8.05   & $95.55\pm0.02$ & $95.32\pm0.03$ & $79.32\pm0.07$  & $78.75\pm0.15$\\
				WideResnet-110 \#2  & 5.97  & $95.60\pm0.03$  & $95.46\pm0.07$  & $79.17\pm0.07$  &  $78.74\pm0.18$\\
				WideResnet-110 \#3  & 4.44   & $95.65\pm0.05$ & $95.30\pm0.02$ & $77.86\pm0.02$ &  $78.84\pm0.13$ \\
				WideResnet-110 \#4  &  3.34 & $95.65\pm0.02$ &  $95.37\pm0.05$ & $78.02\pm0.19$ &  $78.81\pm0.17$ \\
				WideResnet-110 \#5  &  2.53  & $95.47\pm0.07$  & $95.55\pm0.05$ & $77.92\pm0.16$ &  $78.82\pm0.11$\\
				WideResnet-110 Ensemble  &  -   & $95.62\pm0.03$  & $95.86\pm0.08$  & $79.62\pm0.28$ & $80.26\pm0.11$  \\
				%\midrule
				%PreResnet-110 (baseline) & 0  & 94.81 & 94.81 &   &  \\
				%PreResnet-110 \#1  &  32  & & 94.15 &   &   \\
				%PreResnet-110 \#2  &  64 &   &  94.16 &  &   \\
				%PreResnet-110 \#3  &  96  &  &  92.17 &  &  \\
				%PreResnet-110 Ensemble  &  -   &  94.01 &    \textbf{94.79} &  &   \\								
				\hline
		\end{tabular}
	}
	\end{table}
	
\begin{table}[!h]
		\caption{Results of iterative \textbf{$\ell_1$-norm Filters Pruning} \cite{li2016pruning} on  Tiny-Imagenet dataset. The \textit{SLR} column presents the result of pruned networks finetuned with small learning rate while \textit{LLR} column shows the results of same networks finetuned with large learning rate. }
		\label{table:compare_lr_tinyimagenet}
		\centering
		\resizebox{0.5\linewidth}{!}{
			\begin{tabular}{|l|c | cc|}
				\hline
				Model      & \#Param  & SLR & LLR \\
				\hline
				\hline
				Resnet-18 (baseline) & 11.01 & 67.22 & 67.22 \\
				Resnet-18 \#1 & 8.30 & $65.66\pm0.21$ & $66.46\pm0.23$ \\
				Resnet-18 \#2 & 6.17 & $64.04\pm0.17$ & $65.81\pm0.14$ \\
				Resnet-18 \#3 & 4.64 & $63.59\pm0.10$ & $65.37\pm0.07$ \\
				Resnet-18 \#4 & 3.52 & $61.91\pm0.12$ & $64.75\pm0.13$ \\
				Resnet-18 \#5 & 2.71 & $61.06\pm0.32$ & $64.70\pm0.33$ \\
				Resnet-18 Ensemble & - &  $67.63\pm0.21$ & $69.30\pm0.11$ \\
				\hline
				Resnet-34 (baseline) & 21.39 & 68.81 & 68.81 \\
				Resnet-34 \#1 & 15.97 & $68.18\pm0.06$ & $68.66\pm0.18$ \\
				Resnet-34 \#2 & 12.02 & $67.10\pm0.10$ & $68.20\pm0.02$  \\
				Resnet-34 \#3 & 9.13  & $66.41\pm0.07$ & $67.90\pm0.12$ \\
				Resnet-34 \#4 & 6.99 &  $66.05\pm0.19$ &  $67.02\pm0.24$\\
				Resnet-34 \#5 & 5.40 &  $64.93\pm0.14$  &  $67.26\pm0.08$\\
				Resnet-34 Ensemble & - & $69.88\pm0.11$  & $71.31\pm0.13$ \\
				\hline
		\end{tabular}
	}
	\end{table}

\end{appendix}
\end{document}